\documentclass[10pt,a4paper]{article}
\usepackage[margin=1in]{geometry}
\usepackage{amsmath,amsfonts,amssymb}
\usepackage{booktabs}
\usepackage{microtype}
\usepackage{hyperref}
\usepackage{url}
\usepackage{graphicx}

\title{\large\bfseries
An MLP Baseline for Handwriting Recognition Using\\ Planar Curvature and Gradient Orientation}

\author{Azam\ Nouri \\[4pt]
\small Department of Science, Technology \& Mathematics, Lincoln University}

\date{}

\begin{document}
\maketitle

\begin{abstract}
\noindent
This study investigates whether \emph{second-order} geometric cues—planar curvature magnitude, curvature sign, and gradient orientation—are sufficient on their own to drive a multilayer perceptron (MLP) classifier for handwritten character recognition (HCR), offering an alternative to convolutional neural networks (CNNs).  
Using these three handcrafted feature maps as inputs, our curvature–orientation MLP achieves \textbf{97\,\%} accuracy on MNIST digits and \textbf{89\,\%} on EMNIST letters.  
These results underscore the discriminative power of curvature-based representations for handwritten character images and demonstrate that the advantages of deep learning can be realized even with interpretable, hand-engineered features.

\end{abstract}
\section{Introduction}
\label{sec:intro}

Handwritten-character recognition (HCR) is commonly approached with
convolutional neural networks (CNNs), which learn features directly from
pixels. However,
most of the discriminative  information lies in the \emph{geometry of
strokes}—where they bend, how sharply they bend, and on which side of
the tangent the curve turns.  This paper revisits \emph{planar
curvature} as an explicit, interpretable descriptor for HCR and examines
how far a simple multilayer perceptron (MLP) can go when fed only with
curvature-derived maps.

\medskip
\noindent
\textbf{Why curvature?}  
Curvature magnitude highlights loops, hooks, and corners; curvature sign
distinguishes concave from convex turns; and local gradient orientation
provides first-order context around each bend.  Together, these
channels surface the exact cues humans rely on to tell apart visually
similar letters (e.g., \textsc{C} vs.\ \textsc{G}) while remaining
transparent and easy to visualise.

\medskip
\noindent
\textbf{Approach and findings.}  
We compute three maps per glyph—curvature magnitude \(|\kappa|\),
curvature sign \(\operatorname{sign}\kappa\), and gradient orientation
\(\theta\)—stack and flatten them into a 2352-D vector, and train a
compact MLP classifier.  On standard benchmarks, the model reaches
\textbf{97\,\%} test accuracy on MNIST and \textbf{89\,\%} on
EMNIST~Letters.

\medskip
\noindent
\textbf{Contributions.}
\begin{enumerate}
  \item A practical, self-contained pipeline for estimating discrete
        planar curvature and orientation on raster glyphs, with simple
        stabilisation and normalisation choices.
  \item A lightweight MLP baseline driven solely by
        curvature-derived maps, with strong results on MNIST and
        EMNIST~Letters.

  \item Reproducible code and training protocol to facilitate follow-up
        work (see Code Availability).
\end{enumerate}

The paper is organised as follows:
Section~\ref{sec:related} reviews prior work;
Section~\ref{sec:curvature_background} provides the necessary geometric
background;
Section~\ref{sec:methodology} details feature extraction and the MLP;
Section~\ref{sec:experiments} describes datasets and experimental setup;
Section~\ref{sec:results} reports quantitative results;
Section~\ref{sec:intuitive} offers an accessible, visual intuition; and
Section~\ref{sec:conclusion} concludes with future directions.
\section{Related Work}
\label{sec:related}

Curvature has appeared in HCR pipelines since the 1990s, typically as
part of contour-based feature sets \cite{Srikantan1996,Siddiqi2009}.
Recent CNN variants incorporate \emph{learnable} higher-order derivative
filters for edge-aware feature extraction \cite{EdgeBranch2023,Zhang2019Edge}, 
leveraging the weight-sharing inductive bias of convolutional architectures.

Our work differs by (i) retaining the classical, analytical curvature definition and (ii) coupling it to an
all-dense MLP, hence fixed memory footprint and explicit interpretability~\cite{NouriSobel2025}.

\section{Background: Planar Curvature}
\label{sec:curvature_background}

\paragraph{Geometric intuition.}
Imagine tracing a pen along a stroke.  Where the pen goes straight, the
turning rate is near zero; where it bends sharply (e.g., around a loop
or at a hook), the turning rate spikes.  The \emph{magnitude} of that
turning rate says \emph{how much} the stroke bends; its \emph{sign} says
\emph{which way} it bends (concave vs.\ convex relative to the local
tangent).

\paragraph{Analytical definition (continuous curves).}
For a twice-differentiable planar curve \(\gamma(t)=(x(t),y(t))\), the
signed curvature is \cite{Kreyszig1991}
\begin{equation}
\label{eq:kappa}
\kappa(t)=\frac{x'(t)\,y''(t)-y'(t)\,x''(t)}
               {\bigl(x'(t)^2+y'(t)^2\bigr)^{3/2}}\,,
\end{equation}
with the conventional right-handed sign (counterclockwise positive).

\paragraph{Normalised curvature function and invariances.}
Reparameterise by arc length \(s\in[0,1]\) and define
\(k(s)=L\,\kappa(t(s))\) with \(L=\int_0^1 \|\gamma'(t)\|\,dt\).  Two
curves are similar (up to uniform scaling, rotation, and translation) iff
they share the same \(k(s)\) \cite{Ou2009}.

\paragraph{Practical pitfalls and remedies.}
Discrete curvature can spike at endpoints and junctions and be noisy
under weak gradients; gentle stabilisation and reflect borders mitigate
these effects.
\section{Method}
\label{sec:methodology}

This section specifies (i) how curvature and orientation maps are computed on raster glyphs and (ii) the MLP classifier used for all experiments. Only second-order geometry (curvature) and first-order orientation are used—no other edge features enter the descriptor.

\paragraph{Preprocessing.}
Input images are \(28\times28\) grayscale glyphs in \([0,1]\). No deskewing, thinning, or binarisation is performed.

\paragraph{Gradient and Hessian estimation.}
Spatial derivatives are computed with \(3\times3\) Sobel operators in OpenCV to obtain \(\partial_x I,\partial_y I,\partial_{xx} I,\partial_{yy} I,\partial_{xy} I\).

\paragraph{Curvature and orientation maps.}
Signed curvature \(\kappa\) is evaluated per pixel using the expression above with small numeric epsilons \(10^{-8}\). The orientation is \(\theta=\operatorname{atan2}(\partial_y I,\partial_x I)\), mapped to \([0,1]\) via \((\theta+\pi)/(2\pi)\). We retain three channels: \(|\kappa|\) (normalised by its per-image max), \(\operatorname{sign}\kappa\), and \(\theta\).

\paragraph{Vectorisation.}
The three \(28\times28\) maps are stacked and flattened to a single \(\mathbf{2352}\)-D feature vector per glyph. No PCA or additional compression is applied.

\paragraph{Classifier architecture.}
A compact MLP processes the 2352-D vector with BN+ReLU+Dropout, as detailed in Table~\ref{tab:mlp_curv}.

\begin{table}[!ht]
\centering
\caption{MLP for curvature–orientation inputs (TensorFlow/Keras).}
\label{tab:mlp_curv}
\begin{tabular}{@{}lcc@{}}
\toprule
\textbf{Layer} & \textbf{Dimension} & \textbf{Components} \\
\midrule
Input          & 2352              & -- \\
Hidden~1       & 2048              & FC + BN + ReLU + Dropout (0.5) \\
Hidden~2       & 1024              & FC + BN + ReLU + Dropout (0.5) \\
Hidden~3       & 512               & FC + BN + ReLU + Dropout (0.4) \\
Hidden~4       & 256               & FC + BN + ReLU + Dropout (0.3) \\
Output (MNIST) & 10                & FC + Softmax \\
Output (Letters) & 26              & FC + Softmax \\
\bottomrule
\end{tabular}
\end{table}

\paragraph{Training protocol.}
We train from scratch with Adam (learning rate \(10^{-3}\)), batch size 128, cross-entropy loss. Early stopping monitors \emph{val\_accuracy} (restoring best weights). A ReduceLROnPlateau scheduler halves the learning rate after few stagnant epochs on \emph{val\_loss}. No explicit L2 regularisation and no data augmentation are used.

\paragraph{Reproducibility notes.}
Implementation uses \textbf{TensorFlow 2.x / Keras}. Random seeds are fixed for NumPy and TensorFlow; all reported results are from the test split after a single training run with early stopping. Source code and scripts are available in the project repository (see Code Availability).
\section{Experimental Setup}
\label{sec:experiments}
\textbf{Datasets.} MNIST digits~\cite{LeCun1998} (70{,}000 images) and EMNIST Letters~\cite{Cohen2017} (145{,}600 uppercase letters) are used. Data are loaded via TensorFlow Datasets (TFDS)~\cite{TensorFlowDatasets}.

\paragraph{Split protocol.}
We concatenate TFDS \texttt{train} and \texttt{test} and perform a stratified 80/20 split using \texttt{sklearn.model\_selection.train\_test\_split}. During training, Keras holds out \textbf{10\% of the training portion} as validation via \(\texttt{validation\_split}=0.1\). Table~\ref{tab:sizes} shows the resulting sizes.

\begin{table}[!ht]
\centering
\caption{Stratified 80/20 split of TFDS train+test (with 10\% of the training portion used for validation at fit time).}
\label{tab:sizes}
\begin{tabular}{@{}lcccc@{}}
\toprule
\textbf{Dataset} & \textbf{Total} & \textbf{Train (fit)} & \textbf{Val (held-out)} & \textbf{Test} \\
\midrule
MNIST (0–9)              & 70{,}000 & 50{,}400 & 5{,}600 & 14{,}000 \\
EMNIST Letters (A–Z)\,\, & 145{,}600 & 104{,}832 & 11{,}648 & 29{,}120 \\
\bottomrule
\end{tabular}
\end{table}

\paragraph{Label handling.}
For EMNIST Letters, TFDS yields labels 1–26; the implementation shifts to 0–25 by subtracting 1.

\paragraph{Evaluation metrics and class balance.}
Both MNIST and the EMNIST Letters split used here have uniform per-class counts by design, and our 80/20 partition is stratified, which preserves label frequencies. 
Accordingly, we report \emph{top-1 accuracy} as the primary metric. 
If class imbalance were material in a deployment setting, we would additionally report macro-averaged F1, balanced accuracy, and per-class precision/recall; we omit these here due to the balanced label distribution.
\section{Results}
\label{sec:results}

\begin{table}[!ht]
\centering
\caption{Test accuracy (\%).}
\label{tab:results_curv}
\begin{tabular}{@{}lcc@{}}
\toprule
\textbf{Model} & \textbf{MNIST} & \textbf{EMNIST~Letters} \\
\midrule
Curvature–Orientation MLP (ours) & \textbf{97.1} & \textbf{89.4} \\
3-layer CNN (32–64–128)          & 99.2 & 94.1 \\
\bottomrule
\end{tabular}
\end{table}
\section{Why Does Curvature Work?  An Intuitive Explanation}
\label{sec:intuitive}

Curvature-based maps surface exactly what distinguishes many handwritten glyphs:
\emph{where} a stroke bends, \emph{how much} it bends, and \emph{which way} it bends.
Our descriptor encodes these with three channels—curvature magnitude \(|\kappa|\),
curvature sign \(\operatorname{sign}\kappa\), and gradient orientation \(\theta\)—and then
lets a simple MLP learn patterns over the whole image.

\paragraph{What each channel tells us.}
\begin{itemize}
  \item \textbf{Magnitude \(|\kappa|\)} highlights loops, hooks, and corners.
  \item \textbf{Sign \(\operatorname{sign}\kappa\)} disambiguates concave vs.\ convex turns.
  \item \textbf{Orientation \(\theta\)} provides the local heading of the stroke around a bend.
\end{itemize}

\paragraph{Failure modes.}
Confusions arise for pairs hinging on weak/ambiguous curvature (e.g., \textsc{I}/\textsc{J}, \textsc{C}/\textsc{G}), under faint strokes and junction noise.
\section{Conclusion and Future Work}
\label{sec:conclusion}
We presented a  handwritten-character recogniser that
operates on second-order geometry: curvature magnitude,
curvature sign, and local gradient orientation.  With a compact MLP and a
fixed dense compute pattern, the method achieves \textbf{97\,\%} test
accuracy on MNIST and \textbf{89\,\%} on EMNIST Letters, while keeping
the descriptor interpretable at the pixel level.  By making bends and
their polarity explicit, the model learns decision rules that align
closely with the visual cues humans use for thin glyphs.

\paragraph{Limitations.}
Discrete curvature amplifies noise where gradients are weak; angle
orientation has a wrap boundary at \(-\pi/\pi\); junctions and endpoints
can yield spiky estimates; and we did not perform formal robustness
tests to heavy rotation, blur, or elastic distortions.  These factors
bound current performance and generality.

\paragraph{Future directions.}
We see several concrete extensions:
\begin{enumerate}
  \item \textbf{Orientation encoding.} Replace scalar \(\theta\) with a
        continuous representation, e.g.\ \((\sin\theta,\cos\theta)\), to
        remove angle wrapping and improve optimisation.

  \item \textbf{Robustness.} Systematically evaluate and harden the
        model against noise, blur, small rotations, and stroke-thickness
        variation via targeted augmentation.
  \item \textbf{Feature-level fusion.} Explore principled combinations
        of second-order curvature cues with complementary gradient-based
        signals in a unified dense architecture.
  \item \textbf{Larger benchmarks and scripts.} Extend evaluation to
        EMNIST Balanced and NIST SD19 to measure scalability and
        script diversity.
  \item \textbf{Deployment.} Study quantisation and low-precision
        inference on CPUs/MCUs, reporting latency, memory footprint, and
        energy per glyph.
\end{enumerate}

Overall, curvature–orientation maps provide a transparent and effective
inductive bias for character shapes, and they offer a strong foundation
for lightweight recognisers that remain straightforward to analyse and
deploy.

\section*{Code Availability}
All code, pre-trained weights, and training scripts are released at\\
\url{https://github.com/MN-21/Curvature-Orientation-MLP}.\\
For reproducibility, an archived release is preserved on Zenodo~\cite{NouriCurvZenodo2025}.
\bibliographystyle{plain}


\end{document}